\documentclass[runningheads]{llncs}
\usepackage[T1]{fontenc}
\usepackage{graphicx}
\usepackage[hidelinks]{hyperref}
\usepackage{color}

\usepackage[caption=false,font=scriptsize]{subfig}
\usepackage{tikz}
\usepackage{standalone}
\usepackage{times}
\usepackage{epsfig}
\usepackage{graphicx}
\usepackage{amsmath}
\usepackage{amssymb}
\usepackage{booktabs}
\usepackage[separate-uncertainty=true,multi-part-units=single]{siunitx}
\usepackage{mathtools}
\usepackage{array}
\usepackage{multirow}
\usepackage{import}
\usepackage{placeins}
\usepackage[inline]{enumitem}
\usepackage{threeparttable}
\usepackage{pgfplots}
\usepackage{tabularx} 
\usepackage{bm}
\usepackage{layouts}
\usepackage{nicefrac}
\usepackage{pdfpages}
\usepackage{pifont}
\usepackage{subcaption} 
\usepackage[space]{cite}
\makeatletter
\let\NAT@parse\undefined
\makeatother

\definecolor{set2orange}{RGB}{252,141,98}
\definecolor{dark2teal}{RGB}{27,158,119}
\definecolor{chocolate217952}{RGB}{217,95,2}
\definecolor{darkcyan27158119}{RGB}{27,158,119}
\definecolor{lightslategray117112179}{RGB}{117,112,179}
\definecolor{deeppink23141138}{RGB}{231,41,138}
\definecolor{mycyan}{RGB}{0,255,255}
\definecolor{mymagenta}{RGB}{255,0,255}

\newlength\simheight
\pgfplotsset{compat=1.17}

\usetikzlibrary{calc}
\usetikzlibrary{positioning}
\usepgfplotslibrary{groupplots}

\DeclarePairedDelimiterX{\infdivx}[2]{\big(}{\big)}{%
  #1\;\delimsize\|\;#2%
}
\newcommand{\kldiv}{D_{\text{KL}}\infdivx}

\DeclareMathOperator{\E}{\mathbb{E}}

\begin{document}
\title{Quantifying Epistemic Uncertainty\\in Absolute Pose Regression}
%
%
\author{Fereidoon Zangeneh\inst{1,2} \and
Amit Dekel\inst{2} \and \\
Alessandro Pieropan\inst{2} \and
Patric Jensfelt\inst{1}
}

\authorrunning{F. Zangeneh et al.}
%
\institute{KTH Royal Institute of Technology \email{\{fzk,patric\}@kth.se} \and
Univrses AB \email{\{firstname.lastname\}@univrses.com}}
\maketitle              
\begin{abstract}
Visual relocalization is the task of estimating the camera pose given an image it views. Absolute pose regression offers a solution to this task by training a neural network, directly regressing the camera pose from image features. While an attractive solution in terms of memory and compute efficiency, absolute pose regression's predictions are inaccurate and unreliable outside the training domain. In this work, we propose a novel method for quantifying the epistemic uncertainty of an absolute pose regression model by estimating the likelihood of observations within a variational framework. Beyond providing a measure of confidence in predictions, our approach offers a unified model that also handles observation ambiguities, probabilistically localizing the camera in the presence of repetitive structures. Our method outperforms existing approaches in capturing the relation between uncertainty and prediction error.

\end{abstract}
\setcounter{footnote}{0}

\section{Introduction}
Visual localization is the task of estimating the pose of a camera from the image that it captures in the environment. It is a technique used for indoor navigation of robots, augmented reality devices, and autonomous driving \cite{burki2019vizard}. A full visual localization pipeline consists of both frame-to-frame tracking of the camera---or relative pose estimation, as well as global relocalization---or absolute pose estimation. The latter refers to estimating the camera pose from a single image in a previously mapped environment, and is the focus of this work. Visual relocalization has been a long-standing topic of research \cite{se2005vision,schindler2007city,cummins2008fab}, with works in search of map representations and algorithms that can most efficiently and accurately estimate the camera pose. Traditional solutions range from image databases to sparse 3D point models of the world as their map representations, used for image retrieval or keypoint matching to estimate the pose for a novel query image \cite{arandjelovic2016netvlad,jin2021image}. These solutions face a trade-off between efficiency and accuracy. A more recent paradigm for visual relocalization relies on using end-to-end trainable pipelines to directly regress the camera pose from the image by a neural network \cite{kendall2015posenet,shotton2013scene, shavit2021learning,brachmann2021visual}. These end-to-end methods store a map representation in the weights of a neural network, promising higher memory and compute efficiency, as well as better robustness compared to the traditional methods \cite{walch2017image}.

\begin{figure}[t]
    \centering
    \resizebox{\columnwidth}{!}{
        \includegraphics{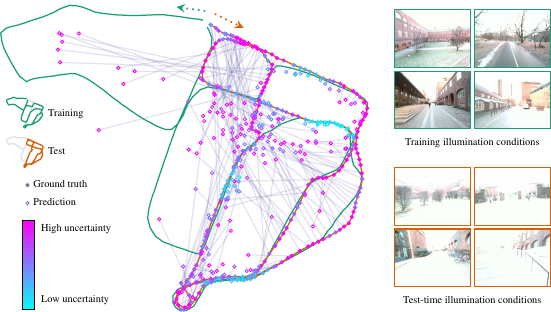}
    }
    \caption{Querying any absolute pose regression network on a trajectory and data domain different from training data results in high prediction errors (depicted by long lines connecting predictions and ground truth). Our proposed epistemic uncertainty quantification approach estimates the likelihood of test samples belonging to the training distribution (visualized by a color map), offering a guide for trusting the predictions. We see that the color of the predictions and their ground truths, which encodes the uncertainty, is highly correlated with the prediction error (the length of the lines).}\label{fig:trajectory}
\end{figure}

Although end-to-end absolute pose regression methods are appealing for their efficiency, they lag behind traditional geometric approaches in terms of accuracy, as evidenced by their performance on visual relocalization benchmarks \cite{toft2020long, sarlin2019coarse}. While the initial motivation for using neural networks in this task was their potential to learn a spatial understanding of the scene, in practice, these models often perform similarly to pose approximations based on image retrieval \cite{sattler2019understanding}, limiting their generalization beyond the training data.  This limitation is further exacerbated by the absence of geometric constraints or verification steps in their prediction process—every input yields a prediction, which can be highly inaccurate. This contrasts with traditional methods, which can identify when there is insufficient evidence to make a reliable prediction. Despite these limitations, the fast inference speed and small memory footprint of pose regression networks has driven significant research efforts aimed at addressing its shortcomings by improving its generalization \cite{moreau2022lens} and accuracy \cite{shavit2021learning, chen2022dfnet}. A remaining challenge in bridging the gap between the current state of absolute pose regression and real-world applications is to accompany network predictions with a reliable measure of confidence. Fig.~\ref{fig:trajectory} illustrates a real-world scenario where a network is tested on a trajectory significantly different from the training/mapped region. The scenario also suffers from strong lighting variations between training and testing, further straining the visual relocalization system. In such cases, the network needs to \textit{know when it does not know}.

Existing research on uncertainty estimation in absolute pose regression has followed two distinct tracks: one focuses on handling ambiguous observations by modeling aleatoric uncertainty \cite{moreau2022coordinet, deng2022deep, zangeneh2023vapor}, while the other---including this paper---aims to quantify epistemic uncertainty to provide confidence in predictions \cite{kendall2016modelling, huang2019prior}. However, the methodologies used in these two tracks differ fundamentally leading to different frameworks. In this work, we build on a solution based on variational autoencoders (VAE) \cite{zangeneh2024cvaepor}, originally proposed to address the challenge of handling ambiguous observations from repetitive structures. We show that VAEs can also be used for quantifying epistemic uncertainty, resulting in a unified framework that can handle both natures of uncertainty. We propose a novel method for quantifying epistemic uncertainty in absolute pose regression to estimate the reliability of the predictions, as shown in Fig.~\ref{fig:trajectory}.

In summary:
\begin{enumerate*} [label=(\arabic*)]
    \item We propose a unified framework for handling epistemic as well as aleatoric uncertainty in absolute pose regression.
    \item We derive a formulation for quantification of epistemic uncertainty in absolute pose regression within a variational framework.
    \item We show how our epistemic uncertainty quantification can estimate the reliability of network predictions.
    \item We perform a thorough evaluation to show that our method outperforms the existing epistemic uncertainty quantification methods.
\end{enumerate*}

\section{Related Work}

\subsection{Visual relocalization}
\subsubsection{Traditional methods}
The well-established solutions to visual relocalization fall into two paradigms: image-based and structure-based methods. Image-based methods store a database of images and their camera poses as a representation of the map. Given a query image, they rely on global image descriptors \cite{arandjelovic2016netvlad} to retrieve the most similar image(s) in the database, hence approximate its pose \cite{torii201524}. The map representation in structure-based methods is a sparse 3D point model of the scene. For localizing a query image, 2D-3D matching is performed between the image and the sparse model, enabling accurate estimation of the camera pose \cite{liu2017efficient, sattler2016efficient}. These two paradigms have distinct strengths and properties on the accuracy-efficiency trade-off, such that they are typically combined in a hierarchical framework for large-scale relocalization \cite{sarlin2019coarse}. Both paradigms, however, exhibit detectable signs of failure when queried with an image that deviates significantly from the map; image-based retrieval computes too little similarity between query and the map, and structure-based matching finds too small inlier sets. So each estimation can be accompanied with a measure of reliability.

\subsubsection{Regression-based methods}
A recent trend in visual relocalization is using end-to-end learning-based pipelines. These methods encode a map representation in the weights of a neural network that directly regresses a geometric quantity from an image. This makes them appealing alternatives to traditional methods for their space and computational efficiency in large scenes, along with data-driven handling of lighting changes, camera blur and texture-less surfaces \cite{walch2017image}. These methods come in two main variants: scene coordinate regression and absolute pose regression. In scene coordinate regression, the neural network regresses the scene's 3D point coordinates from image patches, which are then used in a robust estimation process to compute the camera pose \cite{shotton2013scene}. In absolute pose regression, the neural network directly predicts the camera pose, eliminating the need for the robust estimation step \cite{kendall2015posenet}. Both variants are appealing in their own right. Scene coordinate regression, due to its more local nature of predictions, shows higher generalization capabilities than absolute pose regression, resulting in more accurate predictions \cite{brachmann2021visual}. This is particularly beneficial in presence of occlusions, where scene coordinate regression can leverage image patches that provide consistent information to produce an accurate final pose, while the global pose predicted by an absolute pose regression network may be adversely affected. Scene coordinate regression, however, originally proposed for relocalization with RGB-D images, requires additional, more complex steps for end-to-end training with RGB images \cite{brachmann2018learning, brachmann2021visual}. In contrast, absolute pose regression comes with a simpler training procedure, and despite its shortcomings \cite{sattler2019understanding} remains an attractive alternative. This has promoted research efforts to build upon the original work \cite{kendall2015posenet}, including the development of more effective geometric and photometric loss functions \cite{kendall2017geometric, brahmbhatt2018geometry, chen2021direct}, network architectures \cite{melekhov2017image, walch2017image, wang2020atloc}, extension to multiple scenes \cite{shavit2021learning}, and improved generalization through data synthesis during training \cite{naseer2017deep, ng2021reassessing, moreau2022lens}. Absolute pose regression in its basic form produces a point prediction for any given input through a forward pass of the network, without providing any confidence in the prediction. A complementary line of research centers on incorporating uncertainty estimation into pose regression networks, enhancing their reliability for real-world applications. Our work aligns with this direction.

\subsection{Uncertainty estimation in absolute pose regression}
Uncertainty in neural network predictions arises in two distinct forms: aleatoric uncertainty and epistemic uncertainty \cite{kendall2017uncertainties}. Aleatoric uncertainty arises from inherent noise and ambiguity in the data, which cannot be reduced even with an infinite amount of training data. An example of this in camera relocalization is repetitive structures that appear visually similar but correspond to different camera poses. Epistemic uncertainty, on the other hand, reflects the model's lack of knowledge about a data point---specifically, how well that point fits within the distribution learned during training.

\subsubsection{Modeling aleatoric uncertainty}
Aleatoric uncertainty is typically classified into two types based on its dependence on the input data: heteroscedastic uncertainty, which varies with the input; and homoscedastic uncertainty, which remains constant regardless of the input such as sensor noise. Kendall and Cipola \cite{kendall2017geometric} used the latter paradigm to learn the optimal weighting of translation and orientation error terms during training of each scene. On the other hand, heteroscedastic uncertainty is modeled to account for ambiguities in observations. A common method for modeling heteroscedastic aleatoric uncertainty is by predicting a unimodal parametric distribution \cite{kendall2017uncertainties}. Moreau et al.\ \cite{moreau2022coordinet} applied this technique to absolute pose regression in outdoor environments, integrating it in a tracking loop. To address multimodal posterior distributions caused by repetitive structures, Deng et al.\ \cite{deng2022deep} proposed predicting a mixture model instead of a unimodal distribution. In contrast, Zangeneh et al.\ \cite{zangeneh2023vapor} adopted a non-parametric approach, representing the predicted distribution through samples rather than a mixture model. This approach was later reformulated  with a conditional VAE in~\cite{zangeneh2024cvaepor}, where a VAE is trained to reconstruct camera pose samples conditioned on image observations. As a result of this, the latent space captures the space of pose ambiguities given an observation.

\subsubsection{Quantifying epistemic uncertainty}
A number of works have explored the quantification of epistemic uncertainty in absolute pose regression. The pioneering work of Kendall and Cipolla \cite{kendall2016modelling} follows Bayesian deep learning principles and incorporates dropout layers in the network to approximate variational inference for the posterior distribution of model weights \cite{gal2016dropout}. At test time, Monte Carlo dropout is applied, performing stochastic forward passes over a query image, with the variance of predictions providing a measure of epistemic uncertainty. The use of dropout to generate multiple predictions was later adopted by \cite{huang2019prior}, with the distinction that instead of applying dropout within a Bayesian neural network, it was applied directly to the input image pixels. This approach served a dual purpose of quantifying uncertainty along with enhancing robustness against moving objects, aim at autonomous driving scenarios. More recently, Liu et al.\ \cite{liu2024hr} proposed a functional method for uncertainty quantification, which directly compares the image features and the predicted pose of a query to those of the training samples. The presence of a training sample similar to the query, in both pose and appearance, is interpreted as low epistemic uncertainty for the model, with the intuition that the model must be well-informed on the query.

While conditional VAEs have been previously explored to explicitly model the aleatoric uncertainty of pose regression models \cite{zangeneh2024cvaepor}, they can in theory also be used for likelihood estimation of test samples \cite{kingma2014auto}, and thus implicitly capture model's epistemic uncertainty about the prediction. Specifically, the structured latent space and the agreement between the encoder and decoder that is learned for a training set can be used to detect samples that are unlikely under the training distribution. In this work, we explore this capability of conditional VAEs for epistemic uncertainty quantification in absolute pose regression, assessing its viability as an alternative to existing methods while providing a unified framework to also capture aleatoric uncertainty.

\section{Method}

An ideal visual relocalization solution predicts the true camera pose $y \in \mathrm{SE}(3)$ for any given image $\boldsymbol{x} \in \mathbb{R}^{H \times W \times 3}$ sampled in the scene, where $(\boldsymbol{x}, y) \sim p_\text{true}$. In practice, the real-world distribution $p_\text{true}$ can only be modeled by a finite set of training samples that form an empirical distribution $p_\text{train}$. Successful modeling of $p_\text{train}$ implies generalization to any other unseen empirical distribution $p_\text{test}$ from the underlying $p_\text{true}$. However, there is in practice always a model performance gap between queries from $p_\text{train}$ and $p_\text{test}$. We are interested in a visual relocalization model that once trained, can measure how well a test sample $(\boldsymbol{x}, y) \sim p_\text{test}$ adheres to the modeled distribution $p_\text{train}$, hence estimate how reliable its predictions are. To this end, we frame the visual relocalization task as learning the distribution $p_\text{train}(y \mid \boldsymbol{x})$. Sampling from this conditional distribution performs the task itself \cite{zangeneh2024cvaepor}, while estimating its likelihood for a given sample $(\boldsymbol{x}, y)$ quantifies the degree of its conformity to the modeled distribution $p_\text{train}$. We discuss the training procedure to model and sample from $p_\text{train}(y \mid \boldsymbol{x})$ in Section \ref{sec:cvaepor}. We then lay down our proposed approach to estimate the likelihood of samples and quantify model's epistemic uncertainty in Section \ref{sec:eunq}. We hereafter refer to $p_\text{train}(y \mid \boldsymbol{x})$ as $p(y \mid \boldsymbol{x})$.

\begin{figure*}[t]
    \centering
    \resizebox{\textwidth}{!}{
        \includegraphics{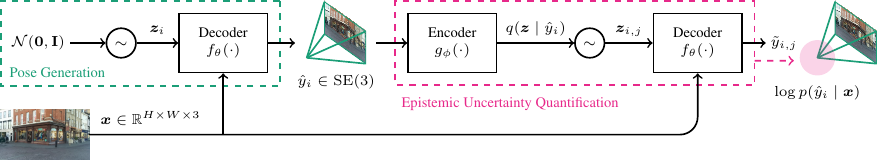}
    }
    \caption{Our pipeline models a scene in a conditional VAE setup. Given a test-time image observation $\boldsymbol{x}$, the decoder is used to sample poses $\hat{y}$ from the posterior distribution of camera poses $p(y \mid \boldsymbol{x})$. Reconstruction of $\hat{y}$ through the VAE pipeline then gives an estimate on $\log p(\hat{y} \mid \boldsymbol{x})$, that is the likelihood of the test sample under the training distribution. This reflects model's epistemic uncertainty about the observation $\boldsymbol{x}$.}\label{fig:pipeline}
\end{figure*}

\subsection{Learning a generative model} \label{sec:cvaepor}
We intend to train a neural network $f_\theta(\cdot)$ that conditioned on a given image $\boldsymbol{x} \in \mathbb{R}^{H \times W \times 3}$ transforms samples $\boldsymbol{z} \in \mathbb{R}^d$ from a noise distribution $p(\boldsymbol{z}) = \mathcal{N}(\mathbf{0}, \mathbf{I})$ to the posterior distribution over camera poses ${y \in \mathrm{SE}(3) \sim p(y \mid \boldsymbol{x})}$. A shown in \cite{zangeneh2024cvaepor}, such a generative network can be trained as the decoder in a conditional VAE pipeline that reconstructs camera poses given images in the scene. We briefly outline this setup and training procedure below, while referring the reader to \cite{zangeneh2024cvaepor} for its design principles. 

\subsubsection{Setup and optimization} 
The conditional VAE consists of an encoder $g_\phi(\cdot)$ and a conditional decoder network $f_\theta(\cdot)$ that are optimized to, together, reconstruct the camera pose $y_i$ for a given image $\boldsymbol{x}_i$ from the training set $(\boldsymbol{x}_i, y_i) \in \mathcal{D}_\text{train}$. The encoder maps each training pose $y_i$ to the mean and covariance of a Gaussian posterior $q(\boldsymbol{z} \mid  y_i)$ that is intended to resemble the true latent posterior distribution $p(\boldsymbol{z} \mid  y_i)$. The decoder, conditioned on the corresponding image $\boldsymbol{x}_i$, then maps samples drawn from this inferred posterior $\boldsymbol{z}_{j} \sim q(\boldsymbol{z} \mid  y_i)$ to reconstructions $\hat{y}_{i,j} \in \mathrm{SE}(3)$ of the original pose $y_i$.

The optimization objective of the pipeline is the evidence lower bound (ELBO) \cite{kingma2019introduction}. From the likelihood of training samples it is possible to derive that
\begin{equation} \label{eq:elbo}
    \begin{split}
        &\log p(y | \boldsymbol{x}) - \overbrace{\kldiv{q(\boldsymbol{z} \mid y)}{p(\boldsymbol{z} \mid y)}}^{\geq 0} \\
        &= \underbrace{\E_{q_\phi(\boldsymbol{z} \mid y)} \log p_\theta(y \mid \boldsymbol{z}, \boldsymbol{x}) - \kldiv{q_\phi(\boldsymbol{z} \mid y)}{p(\boldsymbol{z})}}_{\text{ELBO}}
    \end{split},
\end{equation}
which indicates that maximizing ELBO with optimization variables $\phi$ and $\theta$ over $\mathcal{D}_\text{train}$ optimizes the networks $g_\phi(\cdot)$ and $f_\theta(\cdot)$ that model $p(y \mid \boldsymbol{x})$. In the above expression the subscripts $\phi, \theta$ denote the relation between distributions and weights of the networks they are materialized in. We compute the expected value of the reconstruction likelihood with Monte Carlo samples from $q_\phi(\boldsymbol{z} \mid y)$, following the Gaussian model
\begin{equation} \label{eq:reconstruction}
        \log p_\theta\big(y \mid \boldsymbol{z}, \boldsymbol{x}\big) = -\nicefrac{1}{2}\big(6\log 2\pi + \log \det(\mathbf{\Sigma}) + \boldsymbol{\xi}^T \mathbf{\Sigma}^{-1} \boldsymbol{\xi} \big),
\end{equation}
where a homoscedastic $6 \times 6$ covariance matrix $\mathbf{\Sigma}$ is shared for training all samples and optimized alongside network weights $\theta$ and $\phi$. We take the reconstruction error to be a local correction on the prediction such that $y = \hat{y} \exp(\boldsymbol{\xi}^{\wedge})$, hence define the error vector as $\boldsymbol{\xi} = \log(\hat{y}^{-1}y)^\vee \in \mathbb{R}^6$ in the Lie algebra $\mathfrak{se}(3)$\footnote{$\exp: \mathfrak{se}(3) \mapsto \mathrm{SE}(3)$ is the exponential map from Lie algebra to Lie group and $\log$ is its inverse. The operator $\wedge$ turns $\boldsymbol{\xi}$ into a member of the Lie algebra $\mathfrak{se}(3)$ and $\vee$ is its inverse.}. Defining the error in the tangent space and learning a shared $\mathbf{\Sigma}$ enables automatic adjustment of loss weights across the different degrees of freedom of the camera pose throughout training. This automatic adjustment is inspired by \cite{kendall2017geometric} and eliminates the need for the suboptimal manual tuning performed per dataset \cite{deng2022deep, zangeneh2023vapor, zangeneh2024cvaepor}. However, while \cite{kendall2017geometric} explored learning only two loss weight parameters between translation and rotation error components, we extend this approach to all six degrees of freedom by modeling the full covariance matrix.

\subsubsection{Sample generation}
We can easily sample from $p(y \mid \boldsymbol{x})$ by conditioning the decoder on the image $\boldsymbol{x}$ and passing samples from the prior through the decoder to get $\mathcal{Y} = \{\hat{y}_i = f_\theta(\boldsymbol{z}_i, \boldsymbol{x}) \mid \boldsymbol{z}_i \sim \mathcal{N}(\mathbf{0}, \mathbf{I})\}$. This is illustrated in the left dotted box in Fig. \ref{fig:pipeline}. As discussed in \cite{zangeneh2024cvaepor}, the decoder learns the space of aleatoric pose ambiguities associated with each image, and for ambiguous images, it appropriately splits the latent space such that different latent regions are mapped to distinct camera poses for an image. 

\subsection{Likelihood estimation} \label{sec:eunq}
We propose to quantify the epistemic uncertainty of a trained model about a test sample by computing its marginal likelihood, which we can estimate by importance sampling:
\begin{equation} \label{eq:evidence}
    \begin{split}
        \log p(y &\mid \boldsymbol{x}) = \log \int p_\theta(y \mid \boldsymbol{z}, \boldsymbol{x})\overbrace{p(\boldsymbol{z} \mid \boldsymbol{x})}^{=p(\boldsymbol{z})} d\boldsymbol{z} \\
        &= \log \int q_\phi(\boldsymbol{z} \mid y) \frac{p_\theta(y \mid \boldsymbol{z}, \boldsymbol{x})p(\boldsymbol{z})}{q_\phi(\boldsymbol{z} \mid y)}d\boldsymbol{z} \\
        &= \log \E_{q_\phi(\boldsymbol{z} \mid y)} \frac{p_\theta(y \mid \boldsymbol{z}, \boldsymbol{x})p(\boldsymbol{z})}{q_\phi(\boldsymbol{z} \mid y)} \\
        &\approx \log \frac{1}{M} \sum_{\boldsymbol{z}_j} \frac{p_\theta(y \mid \boldsymbol{z}_j, \boldsymbol{x})p(\boldsymbol{z}_j)}{q_\phi(\boldsymbol{z}_j \mid y)} ~~~ \substack{\boldsymbol{z}_j \sim q_\phi(\boldsymbol{z} \mid y) \\ j=1:M}
    \end{split}
\end{equation}
The equality $p(\boldsymbol{z} \mid \boldsymbol{x}) = p(\boldsymbol{z})$ stems from the assumption that all observations $\boldsymbol{x}$ share the same prior distribution $p(\boldsymbol{z})$ \cite{walker2016uncertain}. At test time, only the image $\boldsymbol{x}$ is available while the true pose $y$ is unknown. Therefore, we propose estimating the expected $\log p(\hat{y}_i | \boldsymbol{x})$ with Monte Carlo generations $\hat{y}_i = f_\theta(\boldsymbol{z}_i, \boldsymbol{x}), \boldsymbol{z}_i \sim \mathcal{N}(\mathbf{0}, \mathbf{I})$. This likelihood measure quantifies the agreement between encoder and decoder when conditioned on $\boldsymbol{x}$, which is how a VAE implicitly captures epistemic uncertainty. The two networks are expected to be in agreement only for in-distribution samples, as the latent space is structured based on the training distribution. In other words, we interpret the likelihood $\log p(\hat{y} | \boldsymbol{x})$ as the network's confidence in its prediction for observation $\boldsymbol{x}$. Our proposed pipeline for pose estimation and epistemic uncertainty quantification is outlined in Fig.~\ref{fig:pipeline}.

\section{Implementation Details} \label{sec:details}

We implement the encoder and decoder networks both as multilayer perceptrons with 5 layers of 512 neurons each, and residual connections between their input and third layers. All neurons (except at output) go through LeakyReLU activations. A pretrained \mbox{ResNet-18} \cite{he2016deep} backbone is used to extract $512$-dimensional conditioning image feature vectors. The input to the decoder is a concatenation of image feature and latent sample vectors. We opt for a $2$-dimensional latent space in all experiments. As the encoder (if expressive enough) is able to learn the surjection of poses to appearances, it is not strictly necessary to condition the encoder on the image features and we obtained similar results for both encoder configurations. We parameterize an $\mathrm{SE}(3)$ element at encoder input and decoder output with a $3$-vector for translation, and a $6$-vector for rotation that can be continuously mapped to a valid rotation matrix by a Gram-Schmidt process \cite{zhou2019continuity}. The $2 \times 2$ latent posterior covariance matrix is parameterized by its log-variances and an $\mathrm{SO(2)}$ rotation. The $6 \times 6$ homoscedastic noise covariance matrix is parameterized by its lower triangular matrix from Cholesky decomposition.

We train the networks for $30\mathrm{k}$ iterations with a learning rate of $1 \text{e}{-4}$ and Adam optimizer with a decoupled weight decay of $1 \text{e}{-2}$ \cite{loshchilov2017decoupled}. We use a batch size of $128$ and draw $100$ Monte Carlo samples to estimate ELBO's expected reconstruction likelihood term in \eqref{eq:elbo}. We found a $0 \rightarrow 1$ warm-up of KL divergence term after $10\mathrm{k}$ iterations to help with better convergence. Following previous works \cite{kendall2015posenet, deng2022deep, zangeneh2024cvaepor} the images are first resized such that the smaller edge is $256$ pixels, then randomly cropped to $244\times244$ squares during training, and center-cropped to the same size at test time. To improve the conditioning of the decoder output, a linear transformation is computed for each scene, allowing the network to predict values within the $[0, 1]$ range for each translation component, which are then mapped to the true metric scale. To enhance robustness to motion blur and lighting variations, we apply random Gaussian blurring with $\sigma \in [0.05, 5]$ during training, along with brightness and color jittering using PyTorch's \cite{paszke2019pytorch} ColorJitter function, with brightness, contrast, and saturation set to $0.2$, and hue to $0.1$.

\section{Experiments}
We aim to assess the efficacy of our proposed likelihood-based quantification of epistemic uncertainty. Defining a measurable ground truth for epistemic uncertainty is inherently challenging. However, we identify and design an evaluation protocol around a key statistical property that the uncertainties should exhibit. We evaluate our method across a variety of operation conditions, comparing it against existing techniques.

\subsection{Evaluation protocol}
Given a trained absolute pose regression model, predictions tend to diverge from the ground truth when test samples deviate from the training distribution. In other words, we expect to observe higher prediction errors for test samples that the model is less informed about and assigns lower likelihood values for. Therefore, the quantified uncertainty should ideally increase with the prediction error. While the exact nature of this relationship remains unknown, we can except one to be a monotonically increasing function of the other. We evaluate this by computing the Spearman's rank correlation coefficient between the estimated negative log-likelihood and prediction error of test samples. This coefficient ranges from $-1$ to $+1$, where $+1$ is the ideal scenario where negative log-likelihood (epistemic uncertainty) is a perfect monotonically increasing function of prediction error. Through this score, we offer a quantitative counterpart to the qualitative uncertainty-error tables and plots used in \cite{huang2019prior, liu2024hr}. As model predictions are made in $\mathrm{SE}(3)$ that does not have a natural metric to measure prediction errors, we report the correlation coefficients for translation and rotation error separately.

\begin{figure*}[t]
    \centering
    \resizebox{\textwidth}{!}{
        \includegraphics{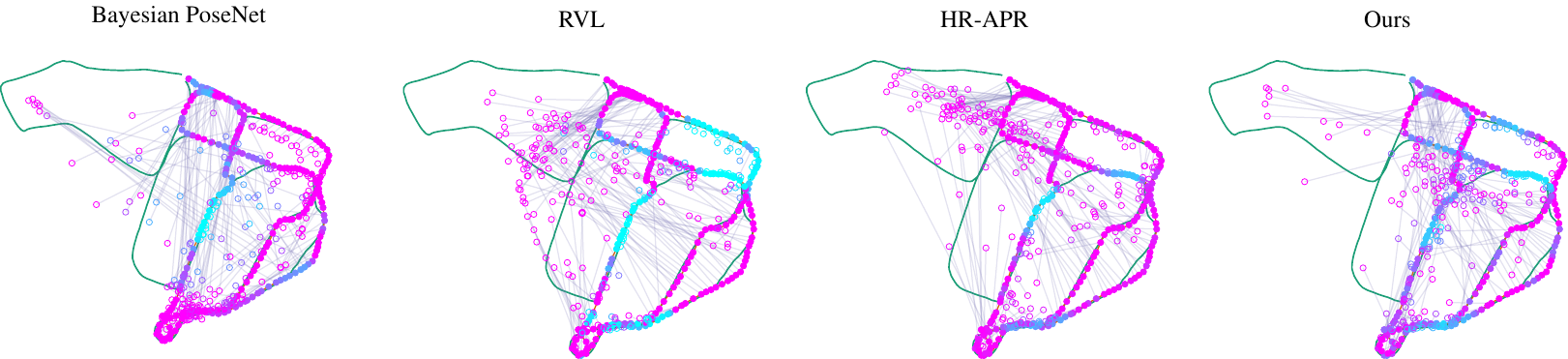}
    }
    \caption{Testing sequence \textcolor{chocolate217952}{\ttfamily\small kth\_day\_09} on models trained on \textcolor{darkcyan27158119}{\ttfamily\small kth\_day\_06} from \cite{nguyen2024mcd}. The lines depict the translation error between the predictions (\textcolor{lightslategray117112179}{$\boldsymbol{\circ}$}) and ground truth (\textcolor{lightslategray117112179}{$\bullet$}), which are colored from \textcolor{mycyan}{cyan} to \textcolor{mymagenta}{magenta} by the epistemic uncertainty. The length of the lines is expected to correlate with epistemic uncertainty, as is the case in our predictions, that is cyan for short lines (small error) to magenta for long lines (large error).
    }\label{fig:all_trajectories}
\end{figure*}

\subsection{Comparison baselines}
We compare our method with existing approaches for epistemic uncertainty estimation in absolute pose regression: Bayesian PoseNet (BPN) \cite{kendall2016modelling}, RVL \cite{huang2019prior}, and HR-APR \cite{liu2017efficient}. For a fair comparison and to rule out the predictive power of different architectures as a factor, we train all methods with the same network architecture and training setup as our own. Given that RVL's dropout statistics are primarily suited for driving scenarios, which we do not evaluate, we use a uniform distribution for their dropout instead. We obtain the distribution of predicted poses for BPN and RVL by $100$ Monte Carlo forward passes. While they propose variance of predictions as the measure of epistemic uncertainty, we found that using negative log-likelihood of the mean prediction benefits both methods in our evaluation metric. So we opt to use that instead to quantify their uncertainties. HR-APR quantifies epistemic uncertainty by a visual dissimilarity measure, which saturates for positionally unlikely samples. 

\subsection{Datasets}
We evaluate our method on the test splits of the real-world datasets Cambridge Landmarks \cite{kendall2015posenet}, 7-Scenes \cite{shotton2013scene}, and ambiguous relocalization sequences of \cite{deng2022deep, zangeneh2023vapor}, covering a variety of outdoor and indoor environments. Additionally, we use two sequences from the Multi-Campus Dataset \cite{nguyen2024mcd} to evaluate the performance of absolute pose regression and uncertainty quantification on test samples far outside the training trajectory.

\begin{table}[t]
    \centering
    \begin{threeparttable}
        \caption{Median prediction error---translation and rotation separately (lower is better)}\label{tab:err}
        \scriptsize
        \setlength{\tabcolsep}{4.0pt}
        \sisetup{detect-weight=true,detect-inline-weight=math}
        \begin{tabular}{@{}lSSSSSSSSSSS@{}}
            \toprule
            \multirow{2.5}{*}{Dataset} & \multicolumn{2}{c}{BPN \cite{kendall2016modelling}} & & \multicolumn{2}{c}{RVL \cite{huang2019prior}} & & \multicolumn{2}{c}{HR-APR \cite{liu2024hr}} & & \multicolumn{2}{c}{Ours}\\
            \cmidrule{2-3}\cmidrule{5-6}\cmidrule{8-9}\cmidrule{11-12}
            & Tra~/~$m$ & Rot~/~$\circ$ && Tra~/~$m$ & Rot~/~$\circ$ && Tra~/~$m$ & Rot~/~$\circ$ && Tra~/~$m$ & Rot~/~$\circ$\\
            \midrule
            Ambiguous \cite{deng2022deep} + Ceiling \cite{zangeneh2023vapor} & 0.94 & 13.13 && 0.87 & 28.15 && 1.36 & 15.47 && \bfseries 0.21 & \bfseries 9.06\\
            7-Scenes \cite{shotton2013scene} & \bfseries 0.19 & \bfseries 9.48 && 0.23 & 10.98 && 0.23 & 11.05 && 0.23 & 11.46\\
            Cambridge Landmarks \cite{kendall2015posenet} & \bfseries 0.67 & 11.61 && 1.23 & 7.69 && 1.29 & 7.38 && 1.21 & \bfseries 7.17 \\
            Multi-Campus: KTH Day \cite{nguyen2024mcd} & \bfseries 22.73 & 61.20 && 29.84 & 36.02 && 23.34 & 43.73 && 24.28 & \bfseries 32.24\\
            \bottomrule
        \end{tabular}
    \end{threeparttable}
    \vspace{-\baselineskip}
\end{table}

\begin{table}[t]
    \centering
    \begin{threeparttable}
        \caption{Spearman's rank correlation coefficient for quantified epistemic uncertainties and prediction errors---translation and rotation separately ($-1$ to $+1$, higher is better)}\label{tab:corr}
        \scriptsize
        \setlength{\tabcolsep}{4.0pt}
        \sisetup{detect-weight=true,detect-inline-weight=math}
        \begin{tabular}{@{}p{2.74cm}SSSSSSSSSSS@{}}
            \toprule
            \multirow{2.5}{*}{Scene} & \multicolumn{2}{c}{BPN \cite{kendall2016modelling}} & & \multicolumn{2}{c}{RVL \cite{huang2019prior}} & & \multicolumn{2}{c}{HR-APR \cite{liu2024hr}} & & \multicolumn{2}{c}{Ours}\\
            \cmidrule{2-3}\cmidrule{5-6}\cmidrule{8-9}\cmidrule{11-12}
            & Tra & Rot & & Tra & Rot & & Tra & Rot & & Tra & Rot\\
            \midrule
            Ceiling & 0.16 & \bfseries 0.68 && 0.18 & 0.59 && 0.25 & 0.01 && \bfseries 0.38 & 0.21 \\
            Blue Chairs & 0.02 & -0.07 && -0.46 & 0.49 && 0.28 & 0.23 && \bfseries 0.64 & \bfseries 0.65\\
            Meeting Table & 0.08 & 0.01 && -0.12 & -0.11 && \bfseries 0.21 & \bfseries 0.20 && 0.16 & -0.01\\
            Seminar & 0.24 & 0.14 && \bfseries 0.82 & \bfseries 0.82 && 0.56 & 0.32 && 0.76 & 0.56 \\
            Staircase & 0.24 & 0.39 && 0.47 & 0.53 && 0.55 & 0.55 && \bfseries 0.67 & \bfseries 0.64\\
            Staircase Extra & 0.10 & 0.20 && 0.19 & \bfseries 0.28 && 0.24 & 0.19 && \bfseries 0.26 & 0.09\\
            \midrule
            Chess & 0.25 & -0.17 && 0.43 & 0.31 && 0.16 & 0.32 && \bfseries 0.43 & \bfseries 0.34\\
            Fire & -0.06 & 0.13 && \bfseries 0.57 & 0.60 && 0.45 & \bfseries 0.62 && 0.38 & 0.53 \\
            Heads & 0.03 & -0.27 && 0.43 & 0.31 && 0.52 & 0.58 && \bfseries 0.70 & \bfseries 0.61\\
            Office & 0.25 & 0.19 && \bfseries 0.50 & 0.45 && 0.48 & \bfseries 0.49 && 0.41 & 0.44\\
            Pumpkin & 0.12 & -0.07 && \bfseries 0.53 & 0.46 && 0.44 & \bfseries 0.47 && 0.50 & 0.43 \\
            Redkitchen & 0.01 & -0.02 && \bfseries 0.55 & \bfseries 0.52 && 0.50 & 0.39 && 0.48 & 0.51 \\
            Stairs & -0.16 & 0.25 && 0.05 & 0.27 && \bfseries 0.44 & 0.30 && 0.39 & \bfseries 0.39\\
            \midrule
            King's College & \bfseries 0.64 & \bfseries 0.76 && 0.12 & 0.41 && 0.12 & 0.09 && 0.06 & -0.17\\
            Old Hospital & 0.27 & \bfseries 0.15 && 0.16 & 0.00 && 0.35 & \bfseries 0.15 && \bfseries 0.52 & 0.12\\
            Shop Fa\c{c}ade & -0.28 & -0.30 && 0.55 & 0.10 && \bfseries 0.58 & 0.08 && 0.37 & \bfseries 0.12\\
            St Mary's Church & 0.15 & 0.15 && 0.21 & 0.24 && \bfseries 0.42 & \bfseries 0.32 && 0.39 & \bfseries 0.32\\
            \midrule
            KTH Day 6 $\rightarrow$ 9 & 0.10 & 0.03 && 0.45 & 0.41 && 0.51 & 0.43 && \bfseries 0.62 & \bfseries 0.52\\
            \specialrule{0.9pt}{2pt}{2pt}
            Average & 0.12 & 0.12 && 0.31 & \bfseries 0.37 && 0.39 & 0.32 && \bfseries 0.45 & 0.35\\
            \bottomrule
        \end{tabular}
    \end{threeparttable}
    \vspace{-\baselineskip}
\end{table}

\section{Results and Discussion}

We start our discussion by examining the behavior of an absolute pose regression network when its test distribution only partially overlaps with the training distribution. This is illustrated for our method in Fig.~\ref{fig:trajectory} and alongside other baselines in Fig.~\ref{fig:all_trajectories}, where the network trained on sequence \texttt{\small kth\_day\_06} is tested on \texttt{\small kth\_day\_09} from Multi-Campus Dataset \cite{nguyen2024mcd}. The test camera trajectory only partially overlaps with the training trajectory considering both position and orientation of the cameras. Additionally, as shown in Fig.~\ref{fig:trajectory}, test images are captured under significantly different illumination conditions. We can see that this distribution shift results in generally large prediction errors across all methods, with median translation errors exceeding $20m$, as reported in Table~\ref{tab:err}. However, Fig.~\ref{fig:trajectory} and \ref{fig:all_trajectories} also highlight that in overlapping stretches of the two trajectories the predictions are better aligned with the ground truth. It is critical for an uncertainty quantification method to distinguish the in- from out-of-distribution regions. Qualitatively, our method shows the highest degree of success at this, with its estimated uncertainty showing the strongest correlation with the prediction error. Next, we quantitatively compare the correlation scores across different methods.

\begin{figure}[!ht]
    \centering
    \resizebox{\textwidth}{!}{
        \includegraphics{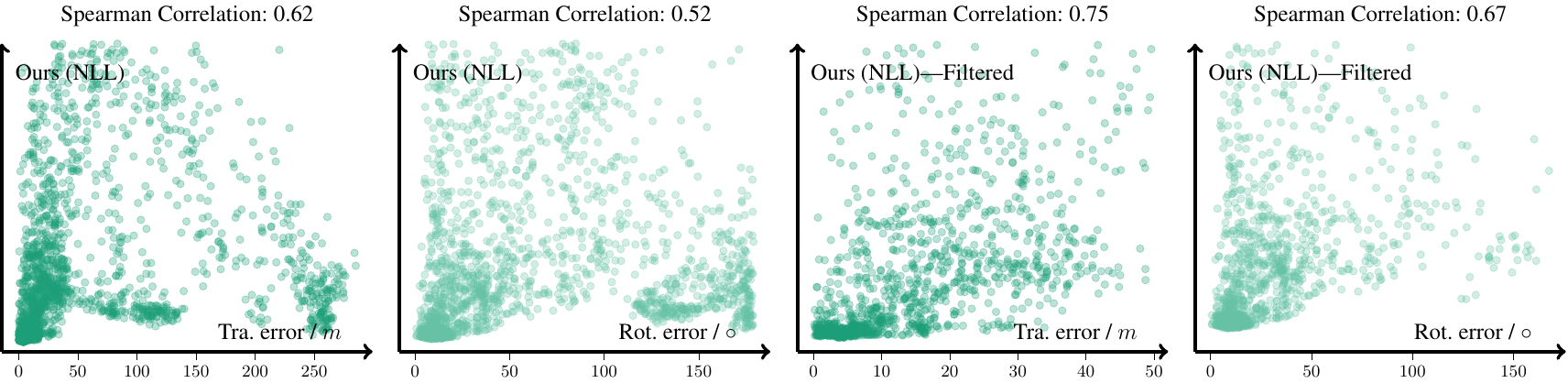}
    }
    
    \resizebox{\textwidth}{!}{
        \includegraphics{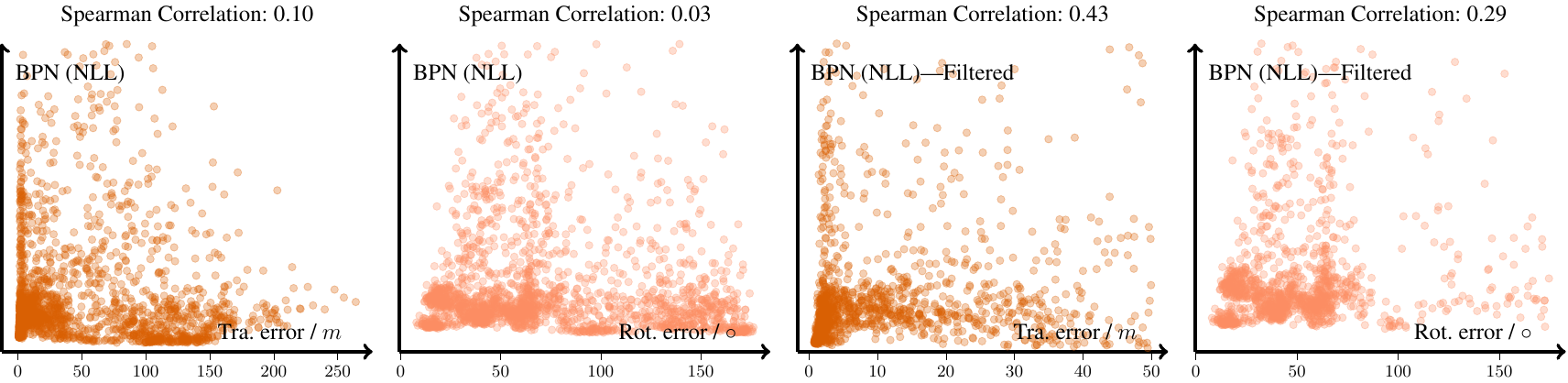}
    }
    
    \resizebox{\textwidth}{!}{
        \includegraphics{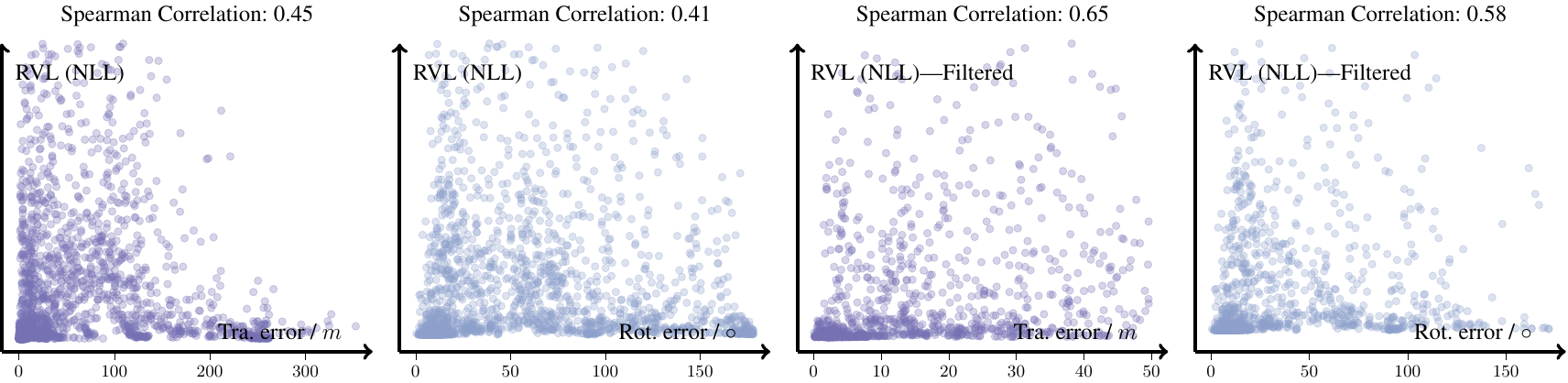}
    }
    
    \resizebox{\textwidth}{!}{
        \includegraphics{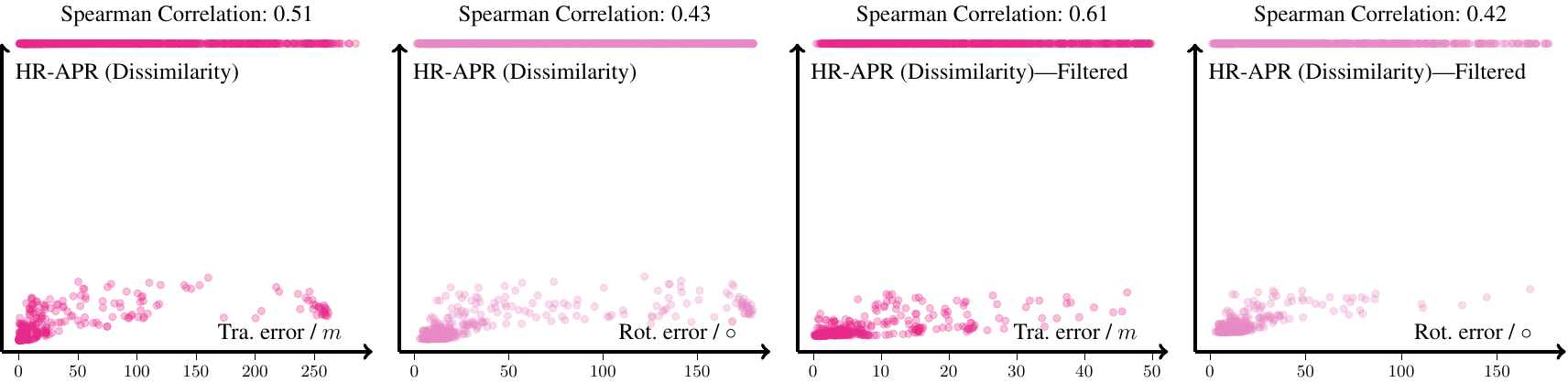}
    }
    \caption{The relationship between quantified epistemic uncertainty and prediction error across methods for our test sequence from Multi-Campus Dataset. The first two columns present scatter plots of all test samples, while the last two columns focus on samples with a translation error of less than $50m$. Since Spearman's rank correlation is invariant to affine changes, we scale and shift the uncertainty distribution in each plot so that their 10th and 90th percentiles align with those of the error distributions for better visualization, hence removing $y$-axis ticks. Our proposed approach demonstrates higher correlations between uncertainty and error, both with and without sample filtering.}\label{fig:scatter}
    \vspace{-\baselineskip}
\end{figure}

Fig.~\ref{fig:scatter} presents the scatter plot of quantified uncertainty against translation and rotation errors for all methods on \texttt{\small kth\_day\_09} test samples. The scattered points should ideally form a monotonically increasing function, corresponding to a Spearman rank correlation of $+1$. Starting from our method, we observe a generally increasing trend in uncertainty for small prediction errors. However, this pattern breaks for larger errors---uncertainty becomes \emph{numb} to increases in error. We hypothesize that this occurs because, once critically far from the training distribution, the VAE model cannot differentiate between different test distributions. In the third and fourth columns of the same figure we filter out data points with translation errors larger than an empirical value of $50m$, which increases correlation coefficients from $0.62$ and $0.52$ to $0.75$ and $0.67$, showing the good performance of our method for test samples are not too far from the training distribution. Turning our attention to the results of other methods in Fig.~\ref{fig:scatter}, we see that other methods show generally lower correlations compared to ours. We also observe the same \emph{numbing} phenomenon in their epistemic uncertainties, though even more severe, as it is harder to empirically filter out their insensitive data points. In the case of HR-APR we can see uncertainty values that saturate at a maximum dissimilarity for predicted poses that deviate significantly from the training samples. We note that while this strategy penalizes the monotonic relationship we evaluate, it can be a practical and cost-effective technique for outlier rejection.

Table~\ref{tab:corr} summarizes the correlation coefficients across different methods and environment. We see that the best performing method can differ between different scenes; BPN performs its best in outdoor scenes, while RVL and HR-APR generally perform better indoors. Our method, however, shows the most consistent performance across all datasets and has the highest average correlation between the quantified epistemic uncertainty and the prediction error. Importantly, our method achieves this while significantly outperforming the baselines methods in ambiguous scenarios with repetitive structures in terms of prediction accuracy, as seen in Table~\ref{tab:err}. In computing the median prediction errors, except for HR-APR that is deterministic, we draw $100$ pose prediction samples per image observation and compute the error for the closest $10\%$ of predicted samples to the ground truth. This is to handle ambiguous scenarios that result in multi-modal posterior distribution of camera poses given an image \cite{zangeneh2024cvaepor}. As evidenced by Table~\ref{tab:err}, on unambiguous datasets our method's accuracy is on par with the baselines.

As an outlook on future work, we reflect on the large variance of correlations across different scenes. Table~\ref{tab:corr} shows that, despite the improvements of our method, some scenes still exhibit correlations close to zero. This highlights the need to enhance the robustness of quantified epistemic uncertainty, potentially by incorporating temporal constraints. Another avenue for future research is addressing the \textit{numbing} effect observed in Fig.\ref{fig:scatter}. One possible approach is to frame this as an out-of-distribution binary classification problem, leveraging a method conceptually similar to HR-APR \cite{liu2017efficient}.

\section{Conclusion}
In this work, we address a key challenge in using absolute pose regression for visual relocalization: estimating the reliability of predictions for queries outside the training distribution. We propose a method for quantifying epistemic uncertainty by estimating the likelihood of observations belonging to the training distribution. We leverage variational autoencoders for modeling this distribution and performing likelihood estimation, which acts as a unified framework for also modeling aleatoric uncertainty to handle observation ambiguities. Through extensive evaluation, we show the effectiveness of our approach in comparison to existing methods, and point out directions for future work.

\begin{credits}
\subsubsection{\ackname} This work was partially supported by the Wallenberg AI, Autonomous Systems and Software Program (WASP) funded by the Knut and Alice Wallenberg Foundation.

\end{credits}
%
%
%
\bibliographystyle{splncs04}
\bibliography{bib}
%




\end{document}